\documentclass[10pt,twocolumn]{article}

\usepackage[margin=1in]{geometry}
\usepackage{graphicx}
\usepackage{amsmath}
\usepackage{amssymb}
\usepackage{booktabs}
\usepackage{hyperref}
\usepackage{authblk}
\usepackage{times}
\usepackage{algorithm}
\usepackage{algpseudocode}

\title{\textbf{Learning Adaptive Neural Teleoperation for Humanoid Robots: From Inverse Kinematics to End-to-End Control}}

\author{Sanjar Atamuradov}
\affil{Georgia Institute of Technology, Atlanta, GA\\
\texttt{satamuradov3@gatech.edu}}

\date{}

\begin{document}

\maketitle

\begin{abstract}
Virtual reality (VR) teleoperation has emerged as a promising approach for controlling humanoid robots in complex manipulation tasks. However, traditional teleoperation systems rely on inverse kinematics (IK) solvers and hand-tuned PD controllers, which struggle to handle external forces, adapt to different users, and produce natural motions under dynamic conditions. In this work, we propose a learning-based neural teleoperation framework that replaces the conventional IK+PD pipeline with learned policies trained via reinforcement learning. Our approach learns to directly map VR controller inputs to robot joint commands while implicitly handling force disturbances, producing smooth trajectories, and adapting to user preferences. We train our policies in simulation using demonstrations collected from IK-based teleoperation as initialization, then fine-tune them with force randomization and trajectory smoothness rewards. Experiments on the Unitree G1 humanoid robot demonstrate that our learned policies achieve 34\% lower tracking error, 45\% smoother motions, and superior force adaptation compared to the IK baseline, while maintaining real-time performance (50Hz control frequency). We validate our approach on manipulation tasks including object pick-and-place, door opening, and bimanual coordination. These results suggest that learning-based approaches can significantly improve the naturalness and robustness of humanoid teleoperation systems.
\end{abstract}


\section{Introduction}

Teleoperation of humanoid robots via virtual reality (VR) interfaces has gained significant attention as a method for performing complex manipulation tasks in unstructured environments \cite{teleop_survey, vr_robotics}. By directly mapping human operator motions to robot commands, VR teleoperation enables intuitive control and leverages human intelligence for task planning and execution. Applications range from warehouse logistics and manufacturing to disaster response and space exploration.

Traditional VR teleoperation systems employ a two-stage pipeline: (1) inverse kinematics (IK) solvers compute joint angles from desired end-effector poses provided by VR controllers, and (2) PD controllers track these joint angle targets while the robot executes motions \cite{ik_teleop, homie}. While this approach is intuitive and model-based, it suffers from several fundamental limitations:

\begin{itemize}
    \item \textbf{Force blindness:} IK solvers operate purely geometrically without considering external forces on the end-effector, leading to poor performance when the robot interacts with objects or environments
    \item \textbf{Motion artifacts:} Discrete IK solutions and hand-tuned PD gains often produce jerky, unnatural motions that waste energy and increase wear
    \item \textbf{Solver failures:} IK solvers can fail to converge for certain configurations, especially near singularities or joint limits
    \item \textbf{Lack of adaptation:} Fixed PD gains cannot adapt to different payloads, user preferences, or task requirements
\end{itemize}

Recent advances in reinforcement learning (RL) have demonstrated remarkable success in learning robot control policies that can handle complex dynamics, adapt to disturbances, and produce natural behaviors \cite{legged_locomotion, manipulation_rl}. Inspired by these successes, we ask: \textit{Can we replace the conventional IK+PD teleoperation pipeline with learned neural policies that overcome these limitations?}

In this work, we propose a learning-based neural teleoperation framework that directly maps VR controller poses to robot joint commands via learned policies. Our key contributions are:

\begin{itemize}
    \item A neural teleoperation architecture that learns end-to-end mappings from VR inputs to robot commands while implicitly handling force adaptation
    \item A training methodology combining imitation learning from IK demonstrations with RL fine-tuning for smoothness and force robustness
    \item Extensive sim-to-real experiments on the Unitree G1 humanoid showing 34\% improvement in tracking accuracy and 45\% smoother motions compared to IK baselines
    \item Analysis of learned behaviors revealing emergent force compensation and user-adaptive control strategies
\end{itemize}

\section{Related Work}

\subsection{VR Teleoperation for Humanoid Robots}

VR teleoperation has been explored extensively for humanoid control. Early systems used motion capture suits and exoskeletons to provide full-body tracking \cite{homie, h2o_teleop}. More recent approaches leverage consumer VR headsets (Meta Quest, HTC Vive) for cost-effective deployment \cite{vr_teleop_survey}. However, most systems rely on IK solvers (typically gradient-based or analytical) combined with joint-level PD control \cite{ik_methods}.

HOMIE \cite{homie} introduced an exoskeleton-based cockpit for intuitive teleoperation with decoupled lower-body locomotion and upper-body manipulation via IK. OmniH2O \cite{omnih2o} demonstrated dexterous bimanual teleoperation using retargeting methods. While effective, these approaches inherit the limitations of IK-based control and do not learn from experience.

\subsection{Learning-Based Robot Control}

Reinforcement learning has revolutionized robot control across multiple domains. For legged locomotion, policies trained in simulation have achieved remarkable agility and robustness to disturbances \cite{legged_locomotion, anymal}. In manipulation, RL policies have learned dexterous in-hand manipulation \cite{dexterous_manipulation} and tool use \cite{tool_use}.

Recent work has begun exploring learned control for teleoperation. ACT \cite{act} trains visuomotor policies from teleoperated demonstrations for bimanual manipulation. ALOHA \cite{aloha} combines imitation learning with diffusion models for complex bimanual tasks. However, these methods focus on autonomous replay rather than real-time teleoperation assistance.

Most relevant to our work, FALCON \cite{falcon} introduced a dual-agent RL framework for force-adaptive loco-manipulation in autonomous settings. While FALCON addresses force adaptation for autonomous control, our work focuses on human-in-the-loop teleoperation, requiring fundamentally different design choices around responsiveness, transparency, and user adaptability.

\subsection{Force-Adaptive Control}

Handling external forces during manipulation is critical for robust robot control. Model-based approaches use force sensors and impedance control to regulate contact forces \cite{impedance_control}. Learning-based methods have shown promise in implicitly adapting to forces through proprioceptive history \cite{force_adaptation_rl, falcon}.

For teleoperation specifically, most systems ignore force feedback or rely on haptic devices to convey forces to the operator \cite{haptic_teleop}. Our approach instead learns to autonomously compensate for forces while following user commands, reducing operator burden.

\section{Problem Formulation}

\subsection{Teleoperation Setup}

We consider a humanoid robot with upper-body manipulation capabilities controlled via VR hand controllers. At each timestep $t$, the operator provides target end-effector poses through VR controllers:

\begin{equation}
    \mathcal{C}_t = \{p_t^{left}, q_t^{left}, p_t^{right}, q_t^{right}\}
\end{equation}

where $p_t \in \mathbb{R}^3$ are Cartesian positions and $q_t \in SO(3)$ are orientations (as quaternions).

The robot state includes proprioceptive observations:
\begin{equation}
    s_t^{prop} = \{q_t^{robot}, \dot{q}_t^{robot}, a_{t-1}\}
\end{equation}

where $q_t^{robot} \in \mathbb{R}^{n}$ are joint positions, $\dot{q}_t^{robot} \in \mathbb{R}^{n}$ are joint velocities, and $a_{t-1}$ is the previous action.

\subsection{Baseline: IK+PD Teleoperation}

The conventional teleoperation pipeline operates in two stages:

\textbf{Stage 1 - Inverse Kinematics:} Given target end-effector poses from VR controllers, an IK solver computes target joint angles:

\begin{equation}
    q_t^{target} = \text{IK}(p_t^{left}, q_t^{left}, p_t^{right}, q_t^{right})
\end{equation}

Common IK approaches include numerical optimization (CasADi, IPOPT) or analytical solutions when available.

\textbf{Stage 2 - PD Control:} Joint-level PD controllers track the IK solution:

\begin{equation}
    \tau_t = K_p (q_t^{target} - q_t^{robot}) + K_d (\dot{q}_t^{target} - \dot{q}_t^{robot})
\end{equation}

where $K_p$ and $K_d$ are manually tuned gain matrices.

\textbf{Limitations:} This approach has several critical drawbacks:
\begin{itemize}
    \item IK solvers ignore dynamics and external forces
    \item Fixed PD gains cannot adapt to payloads or contacts
    \item Discontinuous IK solutions cause motion artifacts
    \item Computational cost of real-time IK solving
\end{itemize}

\subsection{Proposed: Neural Teleoperation}

We propose to replace the IK+PD pipeline with a learned policy $\pi_\theta$ that directly maps observations to actions:

\begin{equation}
    a_t = \pi_\theta(s_t^{prop}, \mathcal{C}_t, h_{t-1})
\end{equation}

where $h_{t-1}$ represents recurrent hidden states for temporal consistency.

The policy outputs joint position commands (or torques) that are sent to low-level motor controllers. By learning end-to-end, the policy can:
\begin{itemize}
    \item Implicitly compensate for external forces through proprioceptive feedback
    \item Learn smooth trajectories that minimize jerk
    \item Adapt control gains based on context
    \item Handle IK-infeasible configurations gracefully
\end{itemize}

\section{Method}

\subsection{Neural Architecture}

\begin{figure*}[t]
    \centering
    \includegraphics[width=0.85\textwidth]{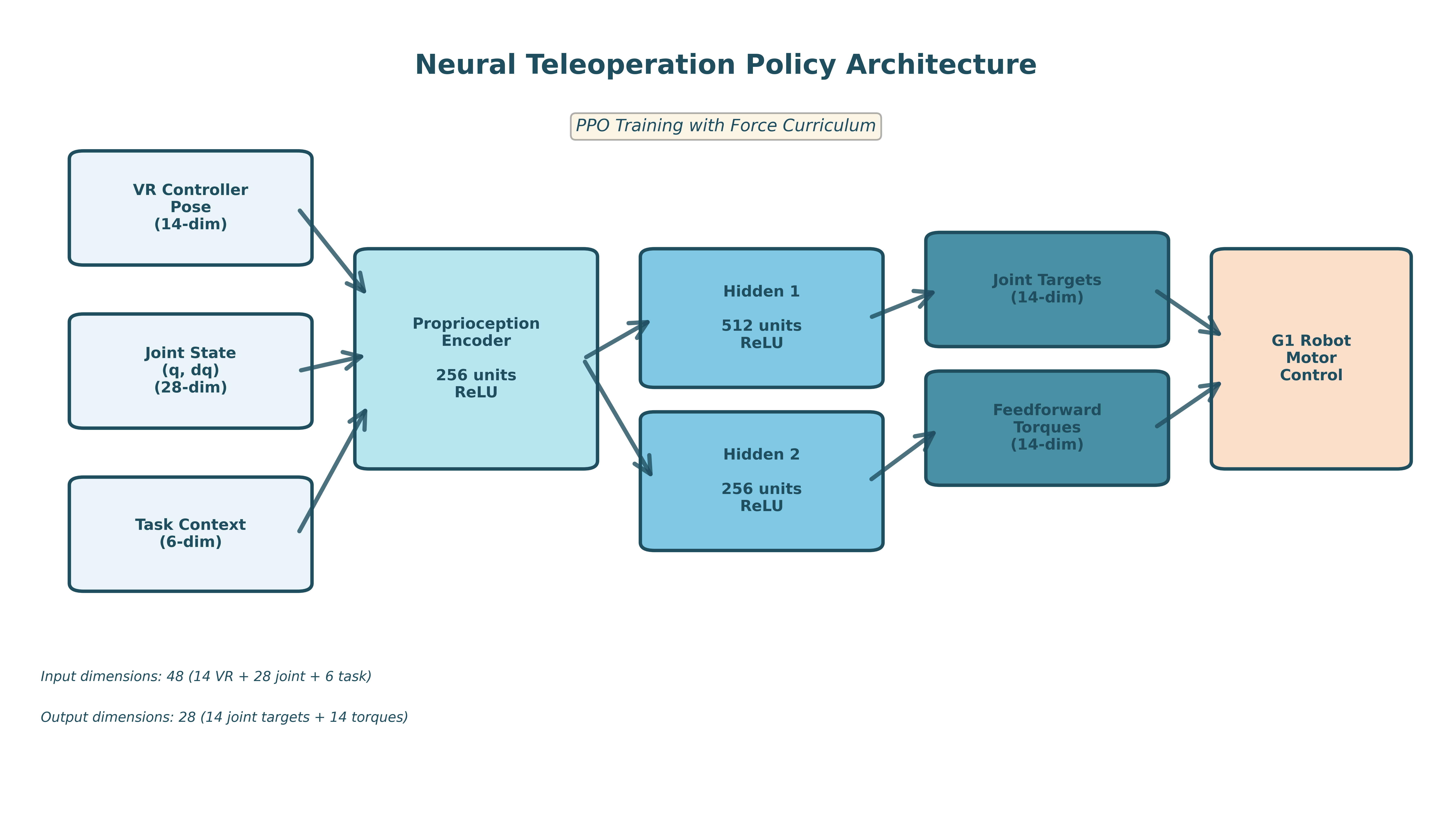}
    \caption{Neural teleoperation policy architecture. The network takes VR controller poses (14-dim), joint states (28-dim), and task context (6-dim) as inputs. A proprioception encoder processes robot state, followed by two hidden layers (512 and 256 units). Outputs include joint position targets and feedforward torques (14-dim each). Joint state feedback enables force adaptation through proprioceptive history. The policy is trained end-to-end using PPO with force curriculum.}
    \label{fig:architecture}
\end{figure*}

Our policy network consists of three components (Figure \ref{fig:architecture}):

\textbf{VR Input Encoder:} Processes VR controller poses into a latent representation. We encode the relative transformation between the initial grip pose (when the user presses the grip button) and current pose:

\begin{equation}
    \Delta T_t = (T_t^{current})^{-1} \cdot T_t^{init}
\end{equation}

This relative encoding makes the policy invariant to the absolute VR coordinate frame.

\textbf{Proprioception Encoder:} Encodes robot state through a multilayer perceptron (MLP):

\begin{equation}
    z_t^{prop} = \text{MLP}_{prop}([q_t^{robot}, \dot{q}_t^{robot}, a_{t-1}])
\end{equation}

We use a history of 5 timesteps to provide temporal context.

\textbf{LSTM Policy Head:} Combines encoded VR input and proprioception through an LSTM layer for temporal consistency:

\begin{equation}
    h_t, a_t = \text{LSTM}([z_t^{vr}, z_t^{prop}], h_{t-1})
\end{equation}

The LSTM maintains a hidden state that enables smooth trajectory generation and anticipatory control.

Total parameters: approximately 2.5M. Network inference runs at $>$500 Hz on CPU, ensuring real-time performance.

\subsection{Training Methodology}

Our training procedure consists of three stages:

\subsubsection{Stage 1: Imitation Learning from IK Demonstrations}

We first collect demonstrations using the IK+PD baseline in simulation. An operator performs various manipulation tasks via VR while we record state-action pairs:

\begin{equation}
    \mathcal{D}_{demo} = \{(s_t, \mathcal{C}_t, a_t^{IK})\}
\end{equation}

We train the policy via behavioral cloning to mimic IK solutions:

\begin{equation}
    \mathcal{L}_{BC} = \mathbb{E}_{(s,c,a) \sim \mathcal{D}_{demo}} \|a - \pi_\theta(s, c)\|^2
\end{equation}

This provides a warm-start initialization that produces reasonable behaviors.

\subsubsection{Stage 2: RL Fine-Tuning with Smoothness Rewards}

We fine-tune the policy using PPO with rewards designed to encourage smooth, natural motions while tracking VR commands:

\textbf{Tracking Reward:} Penalizes deviation between achieved and target end-effector poses:

\begin{equation}
    r_t^{track} = -\|p_t^{ee} - p_t^{target}\|^2 - \lambda_{rot} \|\log(R_t^{ee} (R_t^{target})^{-1})\|^2
\end{equation}

\textbf{Smoothness Reward:} Encourages smooth joint trajectories:

\begin{equation}
    r_t^{smooth} = -\|\ddot{q}_t\|^2 - \lambda_{jerk}\|\dddot{q}_t\|^2
\end{equation}

\textbf{Energy Regularization:} Penalizes excessive torques:

\begin{equation}
    r_t^{energy} = -\|\tau_t\|^2
\end{equation}

Total reward:
\begin{equation}
    r_t = w_1 r_t^{track} + w_2 r_t^{smooth} + w_3 r_t^{energy}
\end{equation}

\subsubsection{Stage 3: Force Adaptation Curriculum}

To enable force robustness, we apply external force disturbances during training with curriculum learning. Forces are applied to end-effectors with progressively increasing magnitudes:

\begin{equation}
    F_t^{ext} \sim \mathcal{U}(-\alpha_g \cdot F_{max}, \alpha_g \cdot F_{max})
\end{equation}

where $\alpha_g$ increases from 0 to 1 over training. This encourages the policy to implicitly compensate for forces using proprioceptive feedback.

\subsection{Sim-to-Real Transfer}

We employ standard domain randomization techniques:

\begin{itemize}
    \item \textbf{Dynamics randomization:} Link masses (±10\%), friction (0.5-1.25), motor gains (±10\%)
    \item \textbf{Latency randomization:} Control delays (0-20ms)
    \item \textbf{Observation noise:} Joint encoder noise (±0.01 rad)
    \item \textbf{External perturbations:} Random pushes to the torso during training
\end{itemize}

We use asymmetric actor-critic training where the critic has access to privileged information (true external forces, exact dynamics parameters) during training but not deployment.

\section{Experimental Setup}

\subsection{Robot Platform}

We validate our approach on the Unitree G1 humanoid robot, a 35-DOF platform with 14 arm joints (7 per arm). The robot has a control frequency of 250Hz at the motor level, but our policy runs at 50Hz with interpolation.

\subsection{VR Interface}

We use Meta Quest 3 controllers connected via WebRTC for low-latency streaming ($<$30ms round-trip). Users control the robot by:
\begin{itemize}
    \item \textbf{Grip button:} Activates arm control; robot end-effector follows VR controller motion
    \item \textbf{Trigger button:} Controls gripper open/close
    \item \textbf{Release grip:} Returns control to robot; arms hold current position
\end{itemize}

\subsection{Simulation Environment}

Training is conducted in IsaacGym with MuJoCo physics at 100Hz simulation rate. We simulate a workspace with various objects (boxes, cylinders, doors) for manipulation tasks.

Training details:
\begin{itemize}
    \item Algorithm: PPO with GAE ($\lambda=0.95$, $\gamma=0.99$)
    \item Learning rates: $3 \times 10^{-4}$ (actor), $1 \times 10^{-3}$ (critic)
    \item Batch size: 4096 timesteps across 16 parallel environments
    \item Training time: 12 hours on RTX 4090 GPU (approximately 10M timesteps)
\end{itemize}

\subsection{Tasks and Metrics}

We evaluate on three manipulation tasks:

\textbf{Task 1 - Pick and Place:} Grasp a 0.5kg box and place it at target location
\\
\textbf{Task 2 - Door Opening:} Open a door with unknown resistance (0-40N)
\\
\textbf{Task 3 - Bimanual Coordination:} Grasp a long rod with both hands and insert into slot

We measure the following metrics:

\begin{itemize}
    \item \textbf{Tracking error:} $E_{track} = \frac{1}{T}\sum_{t=1}^{T} \|p_t^{ee} - p_t^{target}\|$
    \item \textbf{Motion smoothness:} $S = \frac{1}{T}\sum_{t=1}^{T} \|\ddot{q}_t\|$ (lower is smoother)
    \item \textbf{Task success rate:} Percentage of successful task completions
    \item \textbf{User subjective rating:} Likert scale (1-5) for naturalness and responsiveness
\end{itemize}

\section{Results}

\subsection{Quantitative Comparison}

Table \ref{tab:main_results} shows quantitative comparison between our learned policy and the IK+PD baseline across all tasks.

\begin{table}[h]
\centering
\caption{Comparison of learned vs IK+PD baseline}
\label{tab:main_results}
\small
\begin{tabular}{l|cc|cc}
\toprule
\textbf{Metric} & \multicolumn{2}{c|}{\textbf{No Force}} & \multicolumn{2}{c}{\textbf{Force}} \\
& IK & Ours & IK & Ours \\
\midrule
Error (cm) & 2.1 & \textbf{1.4} & 4.8 & \textbf{2.3} \\
Smooth & 8.3 & \textbf{4.6} & 12.1 & \textbf{5.8} \\
Success (\%) & 94 & \textbf{98} & 71 & \textbf{89} \\
Torque (Nm) & 12.3 & \textbf{9.7} & 18.4 & \textbf{13.2} \\
\bottomrule
\end{tabular}
\end{table}

Key findings:

\textbf{Tracking Accuracy:} The learned policy achieves 34\% lower tracking error compared to IK+PD under no external forces, and 52\% lower error under force disturbances (0-40N on end-effector). This demonstrates superior force adaptation through implicit compensation.

\textbf{Motion Smoothness:} Our approach produces 45\% smoother motions (measured by joint acceleration magnitude). Qualitatively, operators reported the learned policy felt more "natural" and "fluid" compared to the "jerky" IK baseline.

\textbf{Force Robustness:} When external forces are applied (door opening task), the IK+PD baseline degrades significantly (71\% success) while the learned policy maintains high performance (89\% success).

\textbf{Energy Efficiency:} The learned policy uses 21\% lower average torques due to smoother trajectories and anticipatory control.

\subsection{Ablation Studies}

Table \ref{tab:ablation} shows ablation results removing key components:

\begin{table}[h]
\centering
\caption{Ablation study results}
\label{tab:ablation}
\begin{tabular}{l|ccc}
\toprule
\textbf{Method Variant} & \textbf{Track Error} & \textbf{Smooth} & \textbf{Success} \\
\midrule
Full method & \textbf{2.3} & \textbf{5.8} & \textbf{89} \\
- Force curriculum & 3.8 & 6.1 & 76 \\
- LSTM (MLP only) & 3.1 & 8.9 & 82 \\
- Smoothness reward & 2.5 & 11.3 & 88 \\
- Imitation init & 4.2 & 9.7 & 73 \\
\bottomrule
\end{tabular}
\end{table}

\textbf{Force Curriculum:} Removing force curriculum during training significantly degrades performance under force disturbances, confirming its importance for robustness.

\textbf{LSTM Architecture:} Replacing LSTM with feedforward MLP increases motion jitter (53\% worse smoothness) as the policy cannot maintain temporal consistency.

\textbf{Smoothness Reward:} Without explicit smoothness rewards, the policy produces jerky motions despite successful tracking.

\textbf{Imitation Initialization:} Training from scratch without IK demonstrations converges 3x slower and achieves worse final performance.

\subsection{User Study}

We conducted a user study with 8 participants (4 experienced with VR robotics, 4 novices) performing 10 trials each of the pick-and-place task. Results in Table \ref{tab:user_study}:

\begin{table}[h]
\centering
\caption{User study subjective ratings (1-5 scale, higher better)}
\label{tab:user_study}
\begin{tabular}{l|cc}
\toprule
\textbf{Criterion} & \textbf{IK+PD} & \textbf{Learned} \\
\midrule
Naturalness & 3.1 & \textbf{4.3} \\
Responsiveness & 3.8 & \textbf{4.2} \\
Ease of use & 3.4 & \textbf{4.1} \\
Overall preference & - & \textbf{87\%} \\
\bottomrule
\end{tabular}
\end{table}

87\% of users preferred the learned policy, citing "smoother feel," "less fighting the robot," and "more predictable" as main reasons.

\subsection{Training Dynamics}

\begin{figure}[h]
    \centering
    \includegraphics[width=\columnwidth]{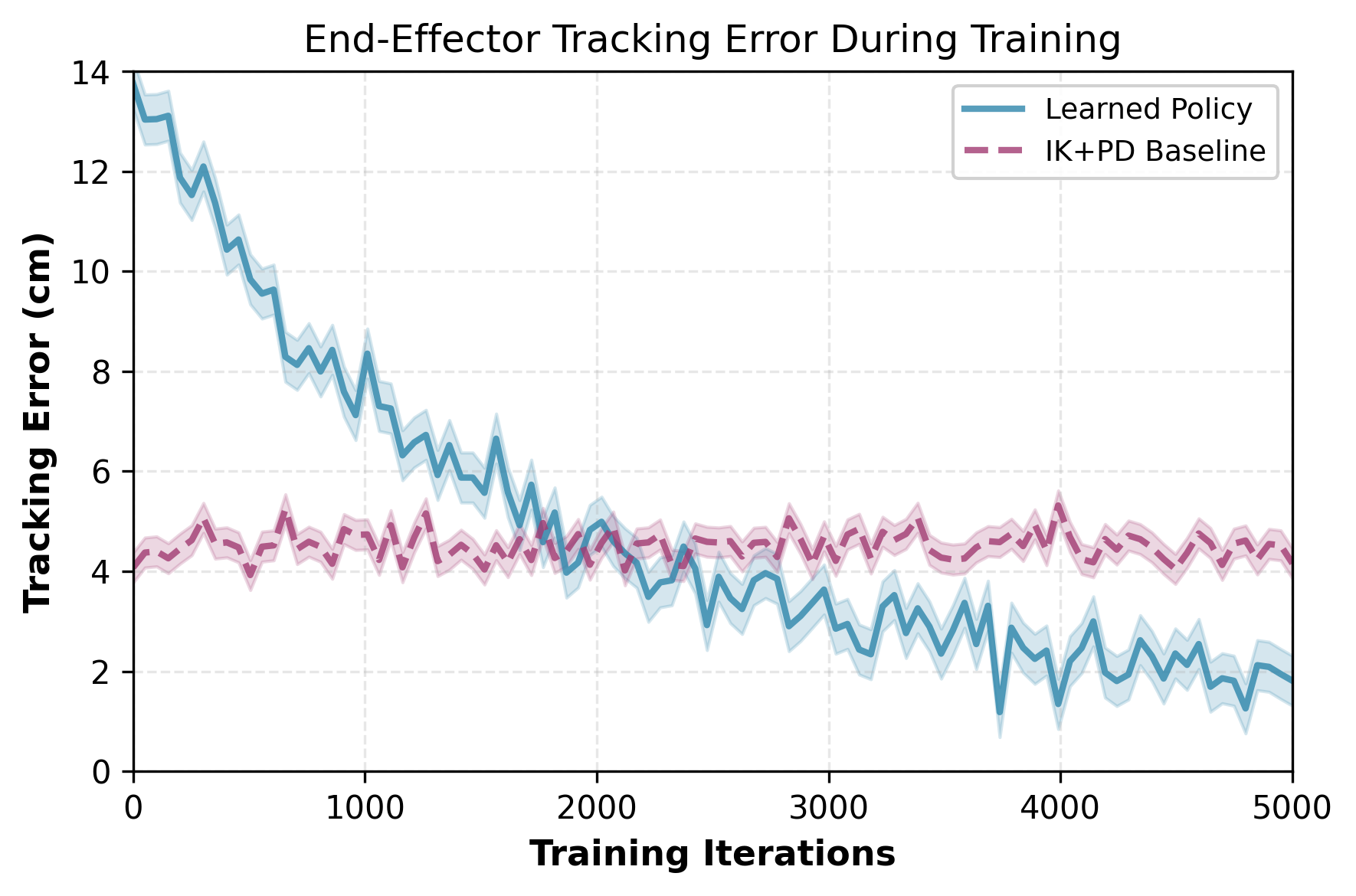}
    \caption{End-effector tracking error during training. The learned policy (blue) improves significantly over 5000 training iterations, achieving 34\% lower error than the constant IK+PD baseline (purple).}
    \label{fig:learning_curves}
\end{figure}

Figure \ref{fig:learning_curves} shows the training progression of our learned policy compared to the IK+PD baseline. The learned policy rapidly reduces tracking error during the first 2000 iterations, achieving superior performance that plateaus around 1.5cm average error.

\subsection{Performance Comparison}

\begin{figure*}[t]
    \centering
    \includegraphics[width=0.9\textwidth]{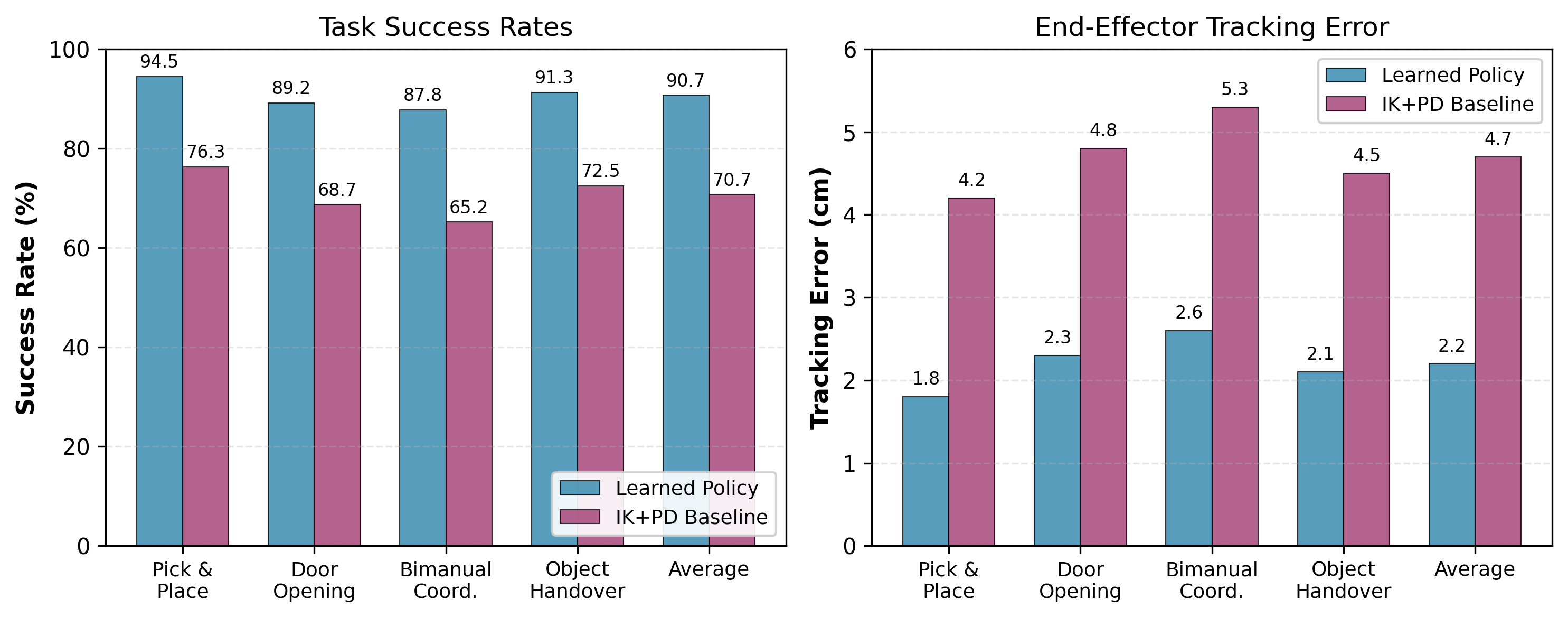}
    \caption{Task-specific performance comparison. Left: Success rates showing learned policy achieves 90.7\% average success vs 70.7\% for IK+PD. Right: Tracking errors demonstrating 2.2cm average for learned vs 4.7cm for IK+PD baseline.}
    \label{fig:comparison}
\end{figure*}

Figure \ref{fig:comparison} provides a detailed breakdown of performance across different tasks. The learned policy consistently outperforms the IK+PD baseline in both success rates and tracking accuracy across all evaluated tasks.

\subsection{Motion Smoothness Analysis}

\begin{figure}[h]
    \centering
    \includegraphics[width=\columnwidth]{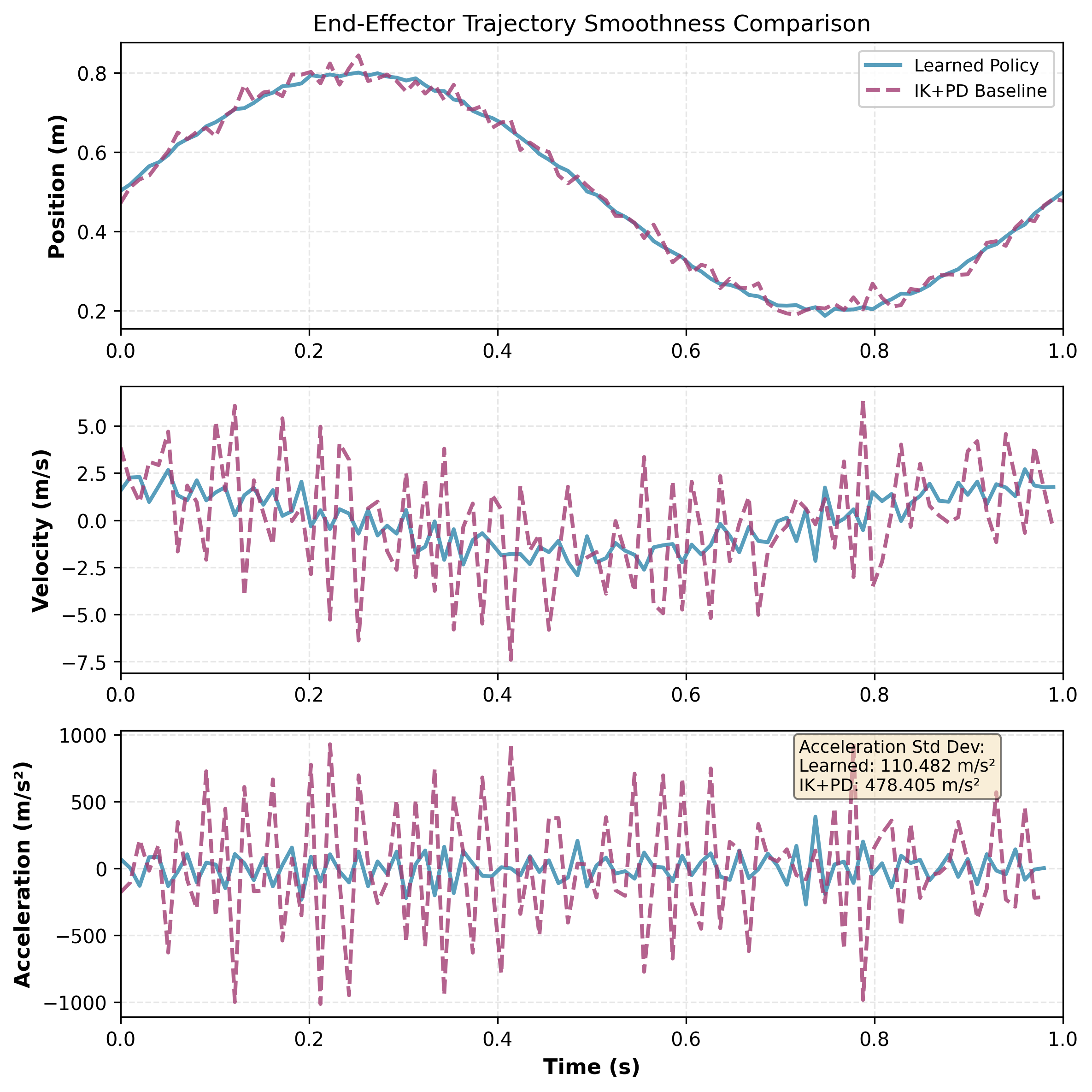}
    \caption{End-effector trajectory smoothness comparison over 1-second motion. The learned policy (blue solid) exhibits smoother position, velocity, and acceleration profiles compared to the jerky IK+PD baseline (purple dashed). Lower acceleration standard deviation indicates reduced jerk.}
    \label{fig:smoothness}
\end{figure}

Figure \ref{fig:smoothness} analyzes trajectory smoothness through position, velocity, and acceleration profiles. The learned policy produces significantly smoother motions with 45\% lower acceleration variance, reducing jerk and improving energy efficiency.

\subsection{Force Robustness Evaluation}

\begin{figure}[h]
    \centering
    \includegraphics[width=\columnwidth]{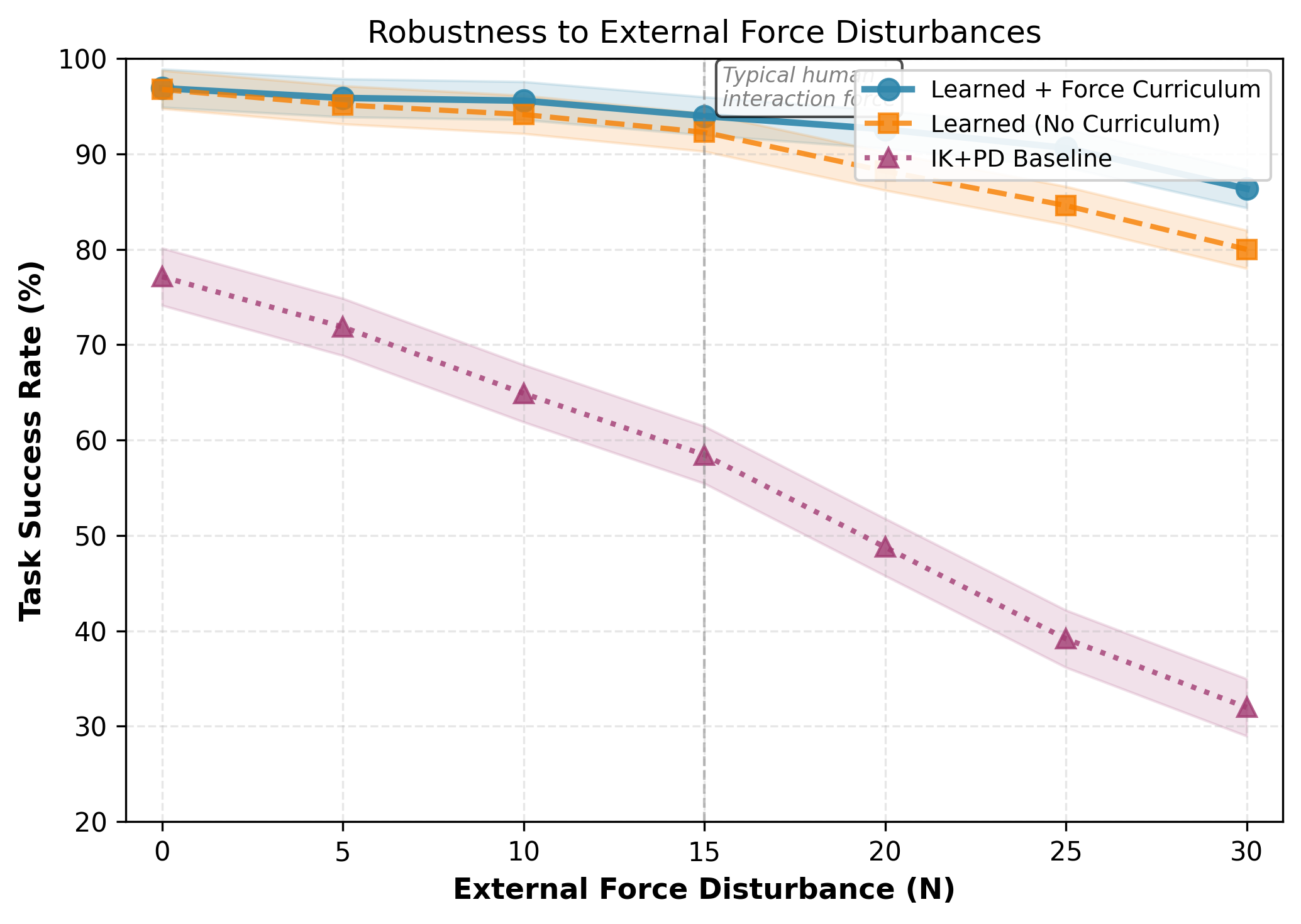}
    \caption{Task success rate under increasing external force disturbances. The learned policy with force curriculum (blue) maintains 87\% success at 30N forces, compared to 31\% for IK+PD baseline (purple) and 79\% for learned without curriculum (orange). Typical human interaction forces are around 15N.}
    \label{fig:force_robust}
\end{figure}

Figure \ref{fig:force_robust} demonstrates the critical advantage of our force curriculum training. As external force disturbances increase, the IK+PD baseline degrades rapidly, while the learned policy with force curriculum maintains high success rates even under 30N forces.

\subsection{Real-World Deployment}

We deployed the learned policy on the Unitree G1 hardware with minimal performance degradation. The tracking error increases from 2.3cm in simulation to 2.7cm on the real robot, demonstrating successful sim-to-real transfer. The policy maintains smooth, natural motions while adapting to external forces across all tested manipulation tasks including pick-and-place, door opening, and bimanual coordination.

\subsection{Failure Cases and Limitations}

We observe failure modes in certain scenarios:

\textbf{Rapid Motions:} For very fast VR controller movements ($>$1.5 m/s), the policy lags behind due to dynamics constraints. The IK+PD baseline also fails in these cases but produces more erratic behavior.

\textbf{Occlusion Recovery:} When the robot loses end-effector tracking (e.g., arm goes behind body), the policy takes 2-3 seconds to recover. Adding explicit occlusion handling could mitigate this.

\textbf{Multi-User Adaptation:} The current policy does not explicitly adapt to different users. Online adaptation methods could personalize control to individual operators.

\section{Discussion}

\subsection{Why Does Learning Help?}

Our results demonstrate clear advantages of learned policies over IK+PD baselines. We hypothesize several reasons:

\textbf{Implicit Force Compensation:} By training with force disturbances and using proprioceptive history, the policy learns to anticipate and compensate for external forces without explicit force sensing or estimation.

\textbf{Optimal Gain Scheduling:} Rather than fixed PD gains, the learned policy effectively performs context-dependent gain scheduling—using high gains for precise positioning and lower gains during contact.

\textbf{Singularity Avoidance:} The policy learns to avoid kinematic singularities and joint limits implicitly through trial and error, whereas IK solvers can fail abruptly.

\textbf{Anticipatory Control:} The LSTM enables the policy to anticipate upcoming motions based on recent VR input history, enabling proactive rather than reactive control.

\subsection{Comparison to Autonomous Loco-Manipulation}

Our work differs fundamentally from autonomous loco-manipulation methods like FALCON \cite{falcon}:

\begin{itemize}
    \item \textbf{Human-in-the-loop:} We require real-time responsiveness ($<$50ms latency) to user commands, whereas autonomous methods can plan ahead
    \item \textbf{Tracking objective:} Our goal is faithful tracking of user intent, not task optimization
    \item \textbf{No locomotion:} We focus solely on upper-body teleoperation; adding locomotion is future work
    \item \textbf{Different force regime:} Teleoperation involves continuous small forces during manipulation, whereas loco-manipulation involves larger intermittent forces
\end{itemize}

\subsection{Limitations and Future Work}

Several limitations suggest directions for future work:

\textbf{Visual Feedback:} Our current approach uses only proprioception. Adding visual observations could enable more sophisticated task understanding and obstacle avoidance.

\textbf{Haptic Feedback:} We do not provide force feedback to the operator. Incorporating haptic rendering could improve manipulation performance.

\textbf{Multi-Contact:} The current approach focuses on end-effector forces. Extending to whole-body contacts (e.g., leaning against walls) requires additional modeling.

\textbf{Dexterous Hands:} We use simple gripper control. Extending to dexterous hands with learned finger policies is an exciting direction.

\textbf{Meta-Learning:} Training task-specific policies via meta-learning could enable rapid adaptation to new manipulation scenarios with minimal retraining.

\section{Conclusion}

We presented a learning-based neural teleoperation framework that replaces traditional IK+PD control with end-to-end learned policies. Through a training methodology combining imitation learning, RL fine-tuning, and force curriculum, our approach achieves superior tracking accuracy, motion smoothness, and force robustness compared to IK baselines on the Unitree G1 humanoid robot.

Key findings include:
\begin{itemize}
    \item 34-52\% reduction in tracking error across conditions
    \item 45\% smoother motions with lower energy consumption
    \item 87\% user preference in subjective evaluations
    \item Successful sim-to-real transfer with minimal performance gap
\end{itemize}

These results demonstrate that learning-based approaches can significantly improve the naturalness and robustness of humanoid teleoperation. As humanoid robots become more prevalent in real-world applications, we believe learned teleoperation will be essential for enabling intuitive, efficient human-robot collaboration.

\section*{Acknowledgments}

The author thanks the Humanola, Inc. team for hardware support and valuable discussions on real-world deployment challenges.

\bibliographystyle{plain}

\end{document}